\documentclass[onecolumn]{fairmeta}

\usepackage[most]{tcolorbox}

\usepackage{amsmath}
\usepackage{amssymb}
\usepackage[utf8]{inputenc} 
\usepackage[T1]{fontenc}    
\usepackage{hyperref}       
\usepackage{url}            
\usepackage{booktabs}       
\usepackage{amsfonts}       
\usepackage{nicefrac}       
\usepackage{microtype}      
\usepackage{xcolor}         
\usepackage{graphicx}
\newcommand{\m}[1]{{A2A}}
\newcommand{\p}[1]{}

\usepackage{wrapfig}
\usepackage{enumitem}
\usepackage{fontawesome5}

\title{MARS Policy: Multimodality Only When It Matters }

\author[1,*,\dagger]{\href{https://jiajindou.github.io/}{\textcolor{black}{Jindou Jia}}}
\author[1,*]{\href{https://morpheus-an.github.io/}{\textcolor{black}{Tuo An}}}
\author[1,*]{Yuxuan Hu}
\author[1]{\href{https://reagan1311.github.io/}{\textcolor{black}{Gen Li}}}
\author[1]{\href{https://lorenzo-0-0.github.io/}{\textcolor{black}{Jingliang Li}}}
\author[1]{Bohan Hou}
\author[1]{\href{https://xyc0212.github.io/}{\textcolor{black}{Xiangyu Chen}}}
\author[1]{Jiaqi Bai}
\author[1]{Bofan Lyu}
\author[1,\dagger]{\href{https://marsyang.site/}{\textcolor{black}{Jianfei Yang}}}
\affiliation[1]{MARS Lab, Nanyang Technological University}

\contribution[*]{Equal contribution}
\contribution[\dagger]{Corresponding authors}

\metadata[\faHome]{\url{https://lorenzo-0-0.github.io/MARS_Policy/}.}
\correspondence{\email{jianfei.yang@ntu.edu.sg}, \email{jindou.jia@ntu.edu.sg}.}

\abstract{
   Imitation learning has become a cornerstone for solving complex robotic manipulation tasks. 
    In particular, multimodality, which enables robots to capture diverse yet valid behavioral patterns, has driven the rapid emergence of generative policies as a dominant paradigm in robot learning.
    However, achieving such multimodality typically relies on stochastic noise initialization and iterative denoising procedures, resulting in substantial training complexity and low inference efficiency. Meanwhile, not all phases of a robotic task inherently require behavioral diversity.
    Motivated by this insight, we propose the \textbf{M}odality-\textbf{A}daptive \textbf{R}obot \textbf{S}ampling (\textbf{MARS}) policy, which adaptively invokes tailored stochasticity only when it is truly beneficial, while reverting to an efficient deterministic learning during single-modal phases. In other words, \textit{the proper amount of noise is injected only at the proper time}. By selectively activating multimodal generation, MARS policy bridges the gap between the multimodal capability of generative policies and the superior training and inference efficiency of deterministic models. Empirical studies across $8$ simulated and $4$ real-world tasks demonstrate that MARS exhibits robust multimodal expressivity and high efficiency, with a $16.67\%$ success rate improvement and an $83.20\%$ inference latency reduction in real-world tests. Counterintuitively, MARS also outpaces deterministic policies in training efficiency on near-deterministic tasks by more effectively modeling nuanced action diversity.
}

\begin{document}

\maketitle

\section{Introduction}

Learning robot policies from expert demonstrations has emerged as a dominant approach for acquiring complex robotic skills. Early deterministic regression policies~\citep{bain1995framework, osa2018algorithmic, jia2025FORESEERR} are efficient in both training and inference, but they often struggle in environments where one observation can correspond to multiple valid actions, a phenomenon known as \textit{multimodality}~\citep{wang2017robust, shafiullah2022behavior, florence2021implicit}. Recent stochastic generative policies, like diffusion policy~\citep{dp,ddpm} and flow matching~\citep{flowmatching}, can generate multimodal actions as conditional generation processes. They introduce stochasticity by sampling from a noise distribution, which allows them to capture diverse plausible actions given the same observation. However, this multimodal capacity comes at the cost of increased training complexity and slower inference~\citep{pan2025much}.

Recent research has begun to address the efficiency bottlenecks of stochastic generative policies. For example, Visual-to-Action (VITA)~\citep{gao2025vita} and Action-to-Action (A2A)~\citep{jia2026action} attempt to streamline the generation process by replacing the standard \textit{Gaussian} noise initialization with informative visual or history action-based priors. By initiating the generation process from more informative starting points, these methods alleviate training complexity and reduce inference steps. However, this grounding in specific priors often compromises the inherent stochasticity, leading to poor multimodal capacity. Concurrently, some approaches utilize consistency models~\citep{li2026ofp, zhang2025flowpolicy, Consistency} or mean-flow~\citep{fang2025omp, geng2026mean} to achieve single-step action generation without compromising stochasticity, but these techniques further introduce additional training overhead, induced by intricate matching objectives.

\begin{figure}
	\centering
	\includegraphics[width=1\linewidth]{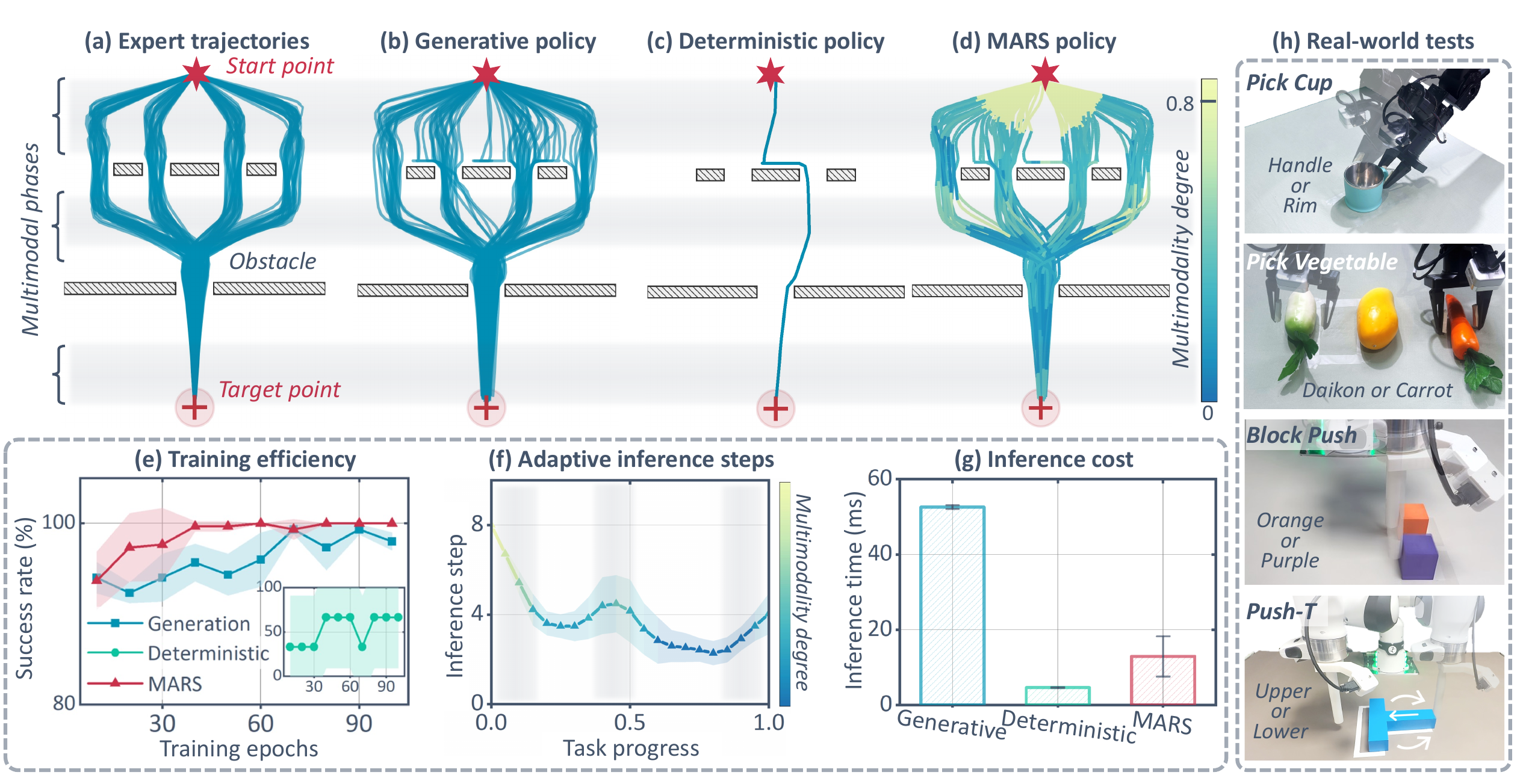}
	\caption{\textbf{Overview of MARS.} \textbf{(a)} Expert trajectories of a 2D navigation task exhibit phase-dependent multimodality, featuring four modes significantly more challenging than Push-T task~\citep{dp,jiang2025streaming}. \textbf{(b)} Stochastic generative policy (e.g., flow matching~\citep{flowmatching}) with standard \textit{Gaussian} noise recovers diverse feasible paths that match the expert multimodal structure. \textbf{(c)} Deterministic policy (e.g., A2A~\citep{jia2026action}) tends to collapse to a single mode and fails to preserve trajectory diversity. Note that deterministic learning can sometimes collide with obstacles due to mode averaging (Fig. \ref{a2a_failed}). \textbf{(d)} The proposed MARS policy maintains multimodality at the proper places while preserving efficient determinism during other phases. The color bar indicates the adaptively identified degree of multimodality, i.e., $\|\mathbf{w}\|_\infty$ defined in Sec.~\ref{Policy_struc}. \textbf{(e)} Compared to the stochastic generative baseline, MARS policy achieves faster convergence by minimizing stochastic injection. More comparisons with baselines, including DDPM~\citep{dp}, BET~\citep{shafiullah2022behavior}, IBC~\citep{florence2021implicit}, Noise-A2A~\citep{jia2026action}, VITA~\citep{gao2025vita}, and ACT~\citep{zhao2023learning}, are provided in Appx.~\ref{appen:2d_benchmark}. \textbf{(f)} MARS policy adaptively schedules inference steps along the task process, resulting in an optimized inference budget \textbf{(g)}. \textbf{(h)} Experimental multimodal tasks: Pick Cup by either its handle or rim, Pick Vegetable targeting either the daikon or carrot, Block Push selecting either the orange or purple object, and Push-T via either the upper or lower path.}
	\label{framework}
    \vspace{-15pt} 
\end{figure}

In practice, the demand for multimodality is often non-uniform across a robotic task. For instance, in the 2D navigation task depicted in Fig.~\ref{framework}(a), certain phases necessitate multimodal capabilities, particularly at junctions, whereas the vast majority of segments follow deterministic and straightforward trajectories. Applying a uniformly high level of stochasticity across all phases unnecessarily complicates the learning process~\citep{pan2025much}, leading to suboptimal performance. Such a phenomenon also exists across a wide range of robotic tasks (Fig.~\ref{framework}(h)). This observation motivates a critical question: can we design a generative policy that \textbf{adaptively modulates its multimodality only when it matters, thereby enhancing learning efficiency during deterministic phases?}

In this work, we propose the \textbf{M}odality-\textbf{A}daptive \textbf{R}obot \textbf{S}ampling (\textbf{MARS}) policy, a novel scheme that can \textit{inject a proper amount of noise only at the proper time}. MARS policy integrates the expressive power of stochastic generative policy~\citep{flowmatching} with the learning efficiency of deterministic regression policy~\citep{jia2026action}. Specifically, we formulate the flow source as a hybrid initial distribution that blends stochastic \textit{Gaussian} noise with history-based deterministic priors. A lightweight \textit{modal scheduling network} is introduced to predict a per-sample weight, dynamically gating the contribution of each source based on the current task context. To capture the underlying multimodal distribution without manual annotations, the network is optimized via a tailored \textit{diversity loss} that directly matches the flow source diversity to the future-action diversity. In this way, MARS policy preserves the multimodality at ambiguous decision points (Fig. \ref{framework}(d)) while retaining efficient training (Fig. \ref{framework}(e)) and fewer-step inference (Fig. \ref{framework}(f-g)) behavior during deterministic phases.

The efficacy of the MARS policy is validated through extensive experiments on $8$ simulated and $4$ real-world tasks, focusing on training and inference efficiency as well as multimodal adaptability. In addition to standard benchmarks including {ManiSkill}~\citep{mu2021maniskill}, {RLBench}~\citep{james2020rlbench}, and {LIBERO}~\citep{liu2023libero}, we also design several specialized tasks with multimodal requirements to evaluate multimodality across directions, grasping poses, velocities, and so on.  Real-world results demonstrate that our proposed method preserves comparative or even better multimodal capabilities while achieving significantly higher learning efficiency, a $83.20\%$ reduction in inference latency, and a $16.67\%$ improvement in average success rate, compared to the flow matching baseline. Notably, even in tasks that appear strategically unimodal yet exhibit nuanced trajectory variations, MARS policy exhibits superior training efficiency over the deterministic one. 


\section{Related work}

\textbf{Multimodal Action Modeling.}
Expert demonstrations in imitation learning are inherently multimodal, where naive mean square error (MSE) regression causes catastrophic \textit{mode averaging}, i.e., yielding geometrically averaged actions that are physically infeasible. To capture high-variance distributions, various frameworks have been developed, such as latent chunk modeling in {ACT}~\citep{wang2017robust,zhao2023learning}, action space discretization in {BET}~\citep{shafiullah2022behavior, lee2024behavior}, and implicit energy estimation in {IBC}~\citep{florence2021implicit}. Beyond these stochastic approaches, deterministic networks with high \textit{Lipschitz} capacity have also been explored for sparse datasets~\citep{pan2025much}. More recently, diffusion policy~\citep{dp, ddpm, song2020score} has become a promising paradigm, as its score-based denoising process offers natural multimodal representation and training stability across complex action distributions.

\textbf{Efficient Action Generation.}
Despite the impressive expressivity of the pioneering diffusion-based generative policy~\citep{dp}, its sub-optimal inference efficiency due to iterative denoising has motivated various subsequent works~\citep{gao2025vita, jia2026action, bai2026flash}. Flow matching~\citep{flowmatching, zhang2025flowpolicy, fang2025omp, li2026ofp} has recently gained significant attention due to its straight-path formulation and improved sampling efficiency, as exemplified by the $\pi$ model family~\citep{intelligence2025pi,intelligence2025pi05, black2024pi_0}. However, the iterative sampling nature still introduces a prohibitive computational barrier~\citep{dp, pan2025much}. To mitigate this, streaming flow policy~\citep{jiang2025streaming} interprets trajectories as continuous flows to facilitate on-the-fly generation, yet it does not fully circumvent the latency inherent in multi-step sampling. In contrast, VITA~\citep{gao2025vita} and A2A~\citep{jia2026action} simplify the generation process by replacing the \textit{Gaussian} noise source with visual- or history-action-based priors, effectively reducing it to a near-deterministic mapping for higher efficiency, though this grounding in specific priors often compromises the inherent multimodal capacity.


\section{Generative \textit{vs.} Deterministic Policies}

Prior to detailing our method, we outline the core principles of stochastic generative and deterministic regression paradigms\footnote{With a slight abuse of terminology, we define a \textit{generative} policy as a stochastic process that recovers target actions from a noise initial distribution, whereas a \textit{deterministic} policy is defined as an unimodal mapping from observations to a single optimal action, optimized via an MSE loss.}. In this work, the prevailing flow matching~\citep{flowmatching, zhang2025flowpolicy, fang2025omp, li2026ofp} serves as the paradigmatic stochastic policy, while A2A~\citep{jia2026action} is employed to represent deterministic policies, a choice motivated by its superior training efficiency and rapid inference performance~\citep{bai2026flash}.

\subsection{Stochastic generative policy}

Flow matching~\citep{flowmatching} defines a continuous-time transport process that maps samples from a source distribution $p_\mathcal{N} = \mathcal{N}(\mathbf{0}, \mathbf{I})$ to the target action distribution $p_\mathcal{T}$ over a time variable $\tau \in [0,1]$. Specifically, given a noise sample $\mathbf{a}_0 \sim p_\mathcal{N}$ and a ground-truth action chunk $\mathbf{a}_1 \sim p_\mathcal{T}$, the interpolated state at time $\tau$ is constructed as $\mathbf{a}_\tau = (1\!-\!\tau)\,\mathbf{a}_0 + \tau\,\mathbf{a}_1$. A neural velocity field $v_\theta(\mathbf{a}_\tau, \tau)$ parameterized by $\theta$ is trained to predict the conditional velocity $\mathbf{a}_1 - \mathbf{a}_0$ at each interpolated point
\begin{equation}
    \mathcal{L}_{\mathrm{fm}} = \mathbb{E}_{\tau \sim \mathcal{U}(0,1),\,\mathbf{a}_0 \sim p_\mathcal{N} ,\,\mathbf{a}_1\sim p_\mathcal{T}}\|v_\theta(\mathbf{a}_\tau, \tau) - (\mathbf{a}_1 - \mathbf{a}_0)\|^2.
    \label{eq:fm_loss}
\end{equation}
At inference, an action is generated by drawing $\mathbf{a}_0 \sim p_\mathcal{N}$ and numerically integrating the learned ODE, i.e., ${d\mathbf{a}_\tau}/{dt} = v_\theta(\mathbf{a}_\tau, \tau)$, from $\tau\!=\!0$ to $\tau\!=\!1$ over discretization steps.

\textbf{What enables multimodality in stochastic generative policies?} In the 2D navigation task depicted in Fig. \ref{framework}, flow matching exhibits robust multimodal capabilities, effectively capturing diverse valid paths. The MSE loss in Eq.~\eqref{eq:fm_loss} actually also averages out the multimodality of the underlying velocity field~\citep{guo2025variational, zhang2025hierarchical}. However, flow matching additionally introduces a stochastic initial distribution to represent the conditional action distribution. Different initial samples, integrated along their respective ODE trajectories, lead to different action predictions. Thus, the multimodal capability of flow matching is enabled by this stochastic initialization, which introduces an additional degree of freedom beyond deterministic observations~\citep{pearce2023imitating}. 

\subsection{Deterministic regression policy}

Unlike standard diffusion or flow matching policies that generate actions from \textit{Gaussian} noise, A2A~\citep{jia2026action} uses proprioceptive historical actions $\mathbf{a}_{0} \sim p_\mathcal{H}$ as the flow starting point for action generation. By replacing the noise-to-action paradigm with an action-to-action formulation, the distributional discrepancy between source and target is significantly reduced, enabling efficient few- or even single-step inference without iterative denoising. 

To facilitate seamless integration with the flow matching later, we reformulate A2A to operate directly in the action space rather than the latent space utilized in the original work~\citep{jia2026action}. In this setting, the learning objective is formalized as 
\begin{equation}
\begin{aligned}
    \mathcal{L}_{\mathrm{A2A}} = \mathbb{E}_{\tau\sim \mathcal{U}(0,1), (\mathbf{a}_0, \mathbf{a}_1) \sim(p_\mathcal{H}, p_\mathcal{T})} \left\| v_\theta(\mathbf{a}_\tau, \tau) - (\mathbf{a}_1 - \mathbf{a}_0) \right\|^2 
    + \lambda_{\text{rec}} \mathbb{E}_{(\mathbf{a}_0, \mathbf{a}_1) \sim(p_\mathcal{H}, p_\mathcal{T})} \left\| \hat{\mathbf{a}}_1 - \mathbf{a}_1 \right\|_1,
    \label{eq:a2a_loss}
\end{aligned}
\end{equation}
where the second $\lambda_{\text{rec}}$ weighted-term represents a \textit{reconstruction loss} based on the ODE-solved action $\hat{\mathbf{a}}_1$. This formulation facilitates fewer-step inference with a predefined fixed-step integration.

Prior research~\citep{dp, pan2025much, jia2026action} has found that generative policies are generally more data-intensive and computationally demanding to train than their deterministic counterparts. Our empirical evaluations across several tasks corroborate this finding. As shown in Fig. \ref{multimodal_sim} and Fig. \ref{singlemodal}, with limited training epochs, the flow matching achieves a lower convergence speed compared to A2A in most cases.

\textbf{What, then, drives the convergence advantage of deterministic policies over stochastic ones?} We attribute this discrepancy primarily to the introduction of stochasticity in generative models, which necessitates more extensive training for the network to effectively internalize the complex mappings from noise-augmented sources to expert actions. To verify this point, we analyze convergence behaviors under different variance settings of the initial distribution, as shown in Fig. \ref{loss_swap}. As the noise variance increases, training loss converges more slowly, and task success rate finally drops.

\section{Modal-Adaptive Robot Sampling Policy}

\begin{wrapfigure}{r}{0.45\textwidth}
    \vspace{-10pt} 
	\centering
	\includegraphics[width=1\linewidth]{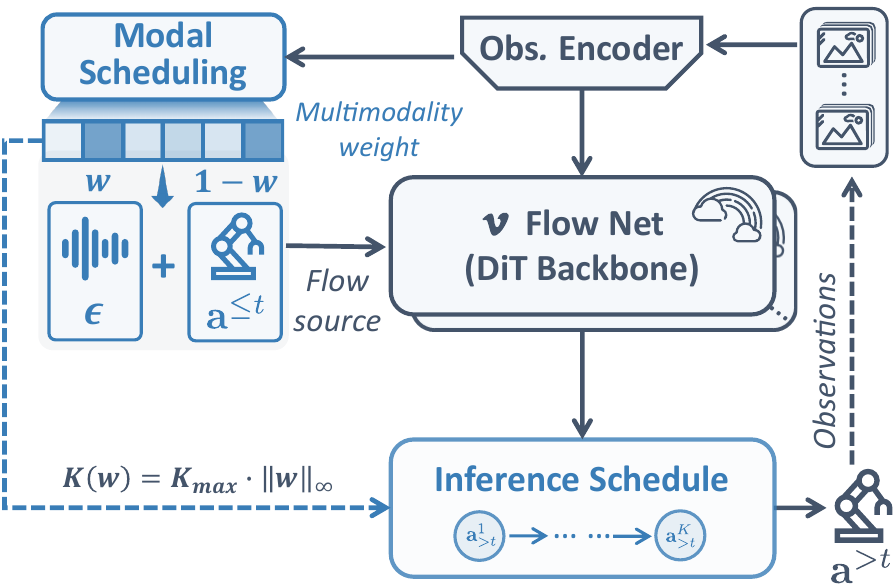}
	\caption{{The architecture of MARS policy.}}
	\label{architecture}
    \vspace{-20pt} 
\end{wrapfigure}

The analysis above reveals that multimodal capability requires stochastic initialization with sufficient variance, yet higher variance degrades learning and inference efficiency. In practice, most robotic tasks contain a mixture of modal phases. Like the 2D navigation task shown in Fig.~\ref{framework}, certain phases require multimodal sampling (\textit{e.g.}, at the junction), while others are inherently unimodal (\textit{e.g.}, at the narrow gate). Applying the same noise level everywhere is suboptimal, as it forces the network to explore unnecessary stochasticity in unimodal phases.

\subsection{Policy structure}
\label{Policy_struc}

To address this gap, we develop the MARS policy, aiming to adaptively adjust stochasticity only when it matters during tasks. Unlike flow matching that always starts from pure \textit{Gaussian} noise, and unlike A2A that starts from history actions, MARS adaptively interpolates between the two extremes based on a learned multimodality weight $\mathbf{w}\in(\mathbf{0},\mathbf{1})^D$, where $D$ is the action dimensionality, $\mathbf{1}$ represents high stochasticity and $\mathbf{0}$ denotes full determinism. The policy structure is illustrated in Fig.~\ref{architecture}.

Specifically, the key module that distinguishes MARS policy from conventional policies is the modal scheduling network. This module is designed to predict a per-dimension weight vector $\mathbf{w}$ to decide multimodality, conditioned on current visual and \textit{optional} proprioceptive observations. Structurally, the modal scheduling network is composed of a {Multi-Layer Perceptron (MLP)} terminated by a {\textit{Sigmoid} activation function} to ensure that the generated weights are constrained within $(\mathbf{0},\mathbf{1})$.

The multimodality weight $\mathbf{w}$ generated from the modal scheduling network directly regulates the composition of the initial flow source. It effectuates a per-dimension convex interpolation between standard \textit{Gaussian} noise and the deterministic historical action prior, i.e.,
\begin{equation}
    \mathbf{a}_0 = (\mathbf{1} - \mathbf{w}) \odot \mathbf{a}^{\leq t} + \mathbf{w} \odot \boldsymbol{\epsilon}, \quad \mathbf{a}^{\leq t}\sim p_\mathcal{H}, \quad \boldsymbol{\epsilon} \sim \mathcal{N}(\mathbf{0}, \mathbf{I}), 
    \label{eq:adaptive_source}
\end{equation}
where $\odot$ denotes the \textit{Hadamard} element-wise product and $\mathbf{a}^{\leq t}$ represents the measured historical action chunk. By operating element-wise, this formulation decouples the multimodal demands across different action dimensions, acknowledging that the requirement for multimodality may vary significantly across different axes within a single task phase. The remaining architecture follows standard flow matching. Visual and proprioceptive inputs are encoded by ResNet-18 for conditioning~\citep{jia2026action}, while the vector field adopts a DiT backbone~\citep{Peebles_2023_ICCV}. 

\subsection{Training objectives}

To coordinate the generative vector field and the modal scheduling network, we optimize a composite objective that balances multimodal expressivity with learning efficiency
\begin{equation}
    \mathcal{L} = \mathcal{L}_{\mathrm{fm}}
               + \lambda_{\mathrm{rec}} \mathcal{L}_{\mathrm{rec}}
                 + \lambda_{\mathrm{div}}\mathcal{L}_{\mathrm{div}},
    \label{eq:total_loss}
\end{equation}
where $\mathcal{L}_{\mathrm{fm}}$ and $\mathcal{L}_{\mathrm{rec}}$ coincide with Eq. \eqref{eq:a2a_loss}, and $\mathcal{L}_{\mathrm{div}}$ represents the \textit{diversity loss}, which will be introduced later. Notably, $\mathcal{L}_{\mathrm{rec}}$ can compromise the multimodal expressivity of the flow matching objective in phases requiring high stochasticity (i.e., as $\mathbf{w} \to \mathbf{1}$), as illustrated in Fig. \ref{fm_reconstruction}. To mitigate this, we implement a per-dimension gating strategy for the reconstruction loss
\begin{equation}
\begin{aligned}
    \mathcal{L}_{\mathrm{rec}} =  \mathbb{E}_{(\mathbf{a}_0, \mathbf{a}_1) \sim(p_\mathcal{H}, p_\mathcal{T})} [\frac{1}{D}(\mathbf{1} - \mathbf{w})^\top | \hat{\mathbf{a}}_1 - \mathbf{a}_1 |],
\end{aligned}
\end{equation}
ensuring that the modality weight adaptively suppresses the single-step \textit{Euler} penalty on dimensions identified as multimodal, thereby preserving distributional diversity when it is most critical. Note that the weight $(\mathbf{1} - \mathbf{w})$ is used solely for loss weighting, with no gradient backpropagation. 

Without additional intervention, the joint optimization of $\mathcal{L}_{\mathrm{fm}}$ and $\mathcal{L}_{\mathrm{rec}}$ naturally biases the flow source toward the deterministic historical prior (i.e., as $\mathbf{w} \to \mathbf{0}$). Because the historical prior resides in closer proximity to the target action than \textit{Gaussian} noise, it provides an easier path for loss minimization, which can lead to a collapse of multimodality. To counteract this tendency, we introduce a diversity loss $\mathcal{L}_{\mathrm{div}}$ to preserve the policy's generative expressivity.

To quantify the degree of multimodality, we introduce \textit{dispersion} $\mathcal{S}$ as a metric of distributional spread. The diversity loss is designed to align the generated source dispersion $\mathcal{S}_{\mathrm{curr}}$ with the target action dispersion $\mathcal{S}_{\mathrm{next}}$, i.e.,
\begin{equation}
\mathcal{L}_{\mathrm{div}} = \mathbb{E} \bigl[\frac{1}{D}\mathbf{1}^\top\mathrm{ReLU}(\mathcal{S}_{\mathrm{next}} - \mathcal{S}_{\mathrm{curr}})\bigr].
\label{eq:diversity_loss}
\end{equation}
To empirically approximate $\mathcal{S}_{\mathrm{next}}$, we evaluate the future action variations among samples sharing similar historical paths. Concretely, we identify the $m$ nearest neighbors $\mathcal{M}(i)$ of sample $i$ within the historical action space using a dataset-wide BallTree~\citep{scikit-learn}. The target spread $\mathcal{S}_{\mathrm{next}}$ and the current source spread $\mathcal{S}_{\mathrm{curr}}$ for sample $i$ are defined as
\begin{equation}
\mathcal{S}_{\mathrm{next}}^{(i)} = \frac{1}{m}\sum_{j\in\mathcal{M}(i)}|\mathbf{a}_{\mathrm{next}}^{(i)} - \mathbf{a}_{\mathrm{next}}^{(j)}|, \quad
\mathcal{S}_{\mathrm{curr}}^{(i)} = \frac{1}{m}\sum_{j\in\mathcal{M}(i)}\bigl|\mathbf{a}_{\mathrm{curr}}^{(i)} - \mathbf{a}_{\mathrm{curr}}^{(j)}\bigr|,
\label{eq:target_spread}
\end{equation}
where $\mathbf{a}_{\mathrm{next}}$ denotes the next action chunk, and $\mathbf{a}_{\mathrm{curr}}$ represents the current action chunk instantiated from the weighted flow sources through Eq. \eqref{eq:adaptive_source}. $\mathcal{S}_{\mathrm{next}}$ is precomputed over the dataset as it relies solely on ground-truth demonstrations. As for $\mathcal{S}_{\mathrm{curr}}$, we reuse updated $\mathbf{w}^{(i)}$ to construct each neighbor's source, so that $\mathbf{w}^{(i)}$ cleanly controls both $\mathbf{a}_{\mathrm{curr}}^{(i)}$ and $\mathbf{a}_{\mathrm{curr}}^{(j)}$.

\textbf{Remark:} This objective penalizes the model only when the source dispersion $\mathcal{S}_{\mathrm{curr}}$ is insufficient to cover the diversity of expert demonstrations. When the current noise level already provides a spread exceeding the target multimodality, the ReLU function ensures that the loss remains inactive. In this regime, the optimization pressure from $\mathcal{L}_{\mathrm{fm}}$ and $\mathcal{L}_{\mathrm{rec}}$ dominates, naturally pulling the flow source back toward the deterministic historical prior, pursuing superior learning efficiency. This interaction establishes a dynamic equilibrium, ensuring that the multimodality weight $\mathbf{w}$ is minimized for learning efficiency while still maintaining the necessary distributional spread required by expert demonstrations.

\subsection{Inference scheduling}

Regarding inference performance, stochastic generative policies suffer from the sampling bottleneck inherent in iterative ODE integration, typically requiring $K \geq 9$ steps to produce high-fidelity actions~\citep{pan2025much,intelligence2025pi,intelligence2025pi05, black2024pi_0}. In contrast, the deterministic policies can operate as a direct one-step mapping~\citep{jia2026action}, resulting in much lower latency. Although generative techniques like consistency models~\citep{li2026ofp, Consistency} or mean-flow~\citep{fang2025omp, geng2026mean} have been proposed to enable single-step generation, they typically introduce additional training complexity. 

In this work, the learned multimodality weight enables adaptive inference scheduling. For samples in predominantly unimodal phases, where the flow source is tightly grounded by the historical prior and lies close to the target, fewer ODE integration steps can yield satisfactory performance. We dynamically determine the per-sample step count based on the maximum weight component across all dimensions, allocating computational budget proportional to the degree of stochasticity, i.e., $K(\mathbf{w}) = K_{\max} \cdot |\mathbf{w}|_\infty $, where $K_{\max}$ is a fixed upper bound on the number of ODE integration steps and $|\mathbf{w}|_\infty$ selects the most multimodal dimension. This scales the inference from a single step for deterministic samples up to the full step budget for highly multimodal ones, ensuring accurate generation without wasted computation.

\section{Evaluation}

\subsection{Experimental setup}
In our experiments, we consider $8$ simulated tasks and $4$ real-world tasks. The simulated tasks consist of four strategically unimodal tasks (\textit{Close Box}, \textit{Stack Cube}, \textit{Pick Cube}, and \textit{Close Drawer}) and four strategically multimodal tasks (\textit{2D Navigation}, \textit{Push Cube}, \textit{Grasp Eyeglasses}, and \textit{Collision Avoidance}). The real-world tasks include \textit{Push-T}, \textit{Pick Cup}, \textit{Block Push}, and \textit{Pick Vegetable}, evaluated on a \textit{Franka Emika Research3} and a \textit{Galaxea R1 Lite} platform. The involved baselines are provided in Appx.~\ref{appen:2d_benchmark} and Appx.~\ref{appen:Unimodal}. Key hyperparameters are standardized across all models to ensure a fair comparison. Training and evaluation settings for different tasks are detailed in Appx. \ref{appen:training}.

We first validate the MARS policy on multimodal tasks, where the MARS policy is expected to exhibit multimodal capability while retaining both learning and inference efficiency, potentially trailing A2A in efficiency but outperforming flow matching. Afterwards, we implement MARS on unimodal tasks to seek additional compelling findings.

\begin{figure}
	\centering
	\includegraphics[trim={0 18 0 0}, width=1\linewidth]{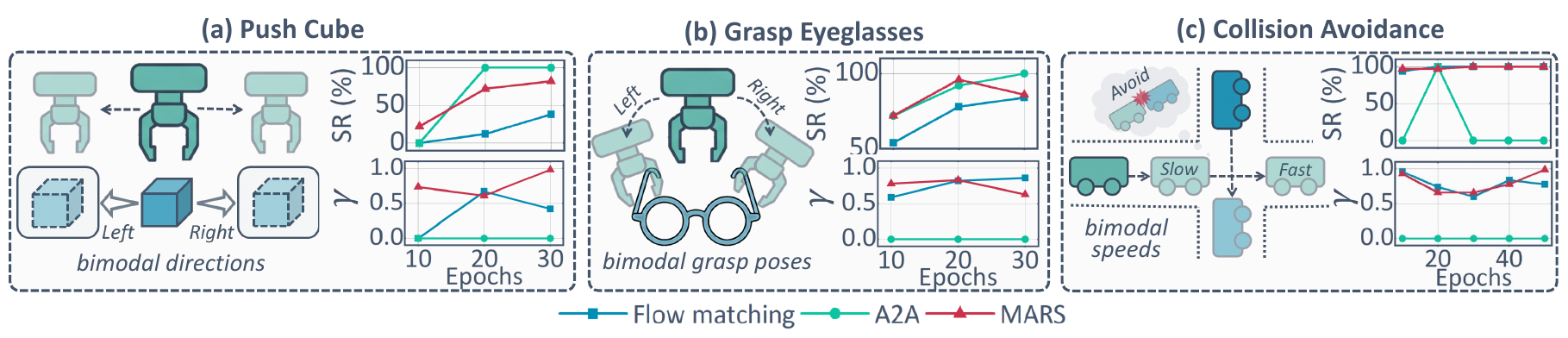}
	\caption{\textbf{Learning performance on strategically multimodal tasks.} (a) Push Cube with bimodal trajectories: the gripper pushes the cube either left or right with equal validity. (b) Grasp Eyeglasses with bimodal grasp poses: the gripper grasps the eyeglasses from either the left or right side. (c) Collision Avoidance with bimodal speeds: the agent may either slow down or accelerate to safely cross the intersection. Each benchmark is accompanied by its training efficiency (\textit{upper panel}: Success Rate, SR) and \textit{modal balance} metric (\textit{lower panel}: $\gamma$) over training epochs.}
	\label{multimodal_sim}
\end{figure}

\subsection{Multimodal evaluation}

\textbf{Simulation results.} Fig.~\ref{framework} and Fig.~\ref{multimodal_sim} summarize the learning efficiency and multimodal characteristics of different algorithms on the 4 simulational multimodal tasks. Focusing first on training efficiency, as shown in Fig.~\ref{framework}(e) and Fig.~\ref{multimodal_sim}, the MARS policy continues to converge faster than flow matching. In contrast, A2A suffers from the mode averaging issue and exhibits highly unstable success rates, occasionally converging the fastest, but often failing entirely on the same task.
To further quantify multimodal capability, we adopt a \textit{modal balance} metric $\gamma \in [0, 1]$, where values closer to 1 denote more balanced mode coverage (Appx. \ref{appen:metrics}).

As shown in the modal curve $\gamma$ of Fig.~\ref{multimodal_sim}, MARS policy retains a modal balance comparable to that of flow matching, demonstrating that the learning efficiency does not come at the cost of multimodal expressivity.
An interesting observation arises from the color bar in Fig.~\ref{framework}(d), that is, preserving full multimodal capability does not always require \textit{Gaussian} noise with unit variance; often a smaller variance suffices. This highlights its noise-adaptivity: \textit{no more stochasticity than necessary}.

Regarding inference efficiency, we further plot the inference step count and cost time curves on the 2D Navigation task in Figs.~\ref{framework}(f) and (g). Benefited from our adaptive multimodal scheduling, the inference step count varies on the fly. More steps are allocated to regions of high stochasticity, while near-deterministic regions are resolved with fewer steps, leading to an optimized inference budget.

\begin{figure}
	\centering
	\includegraphics[trim={0 24 0 0},width=1\linewidth]{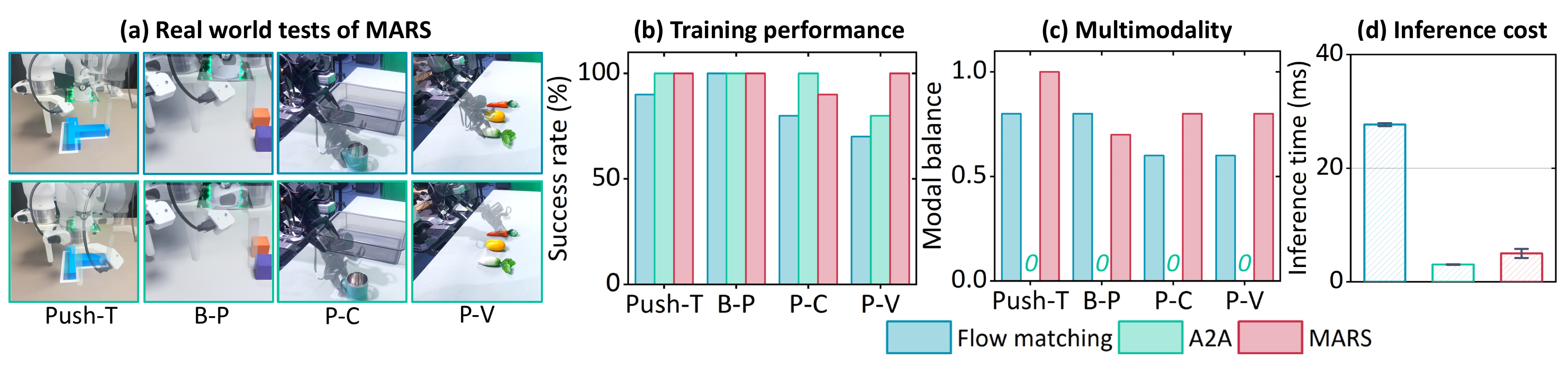}
	\caption{\textbf{Real-world experimental results.} 
    (a) Real-world deployment results on the {Push-T}, Block Push (B-P), {Pick Cube} (P-C), and Pick Vegetable (P-V) tasks. See Fig.~\ref{pusht}-\ref{pickcup} for detailed snapshots.
    (b) Success rate comparison. 
    (c) Multimodality evaluation quantified by the modal balance metric $\gamma$.
    (d) Inference cost analysis.}
	\label{realtest}
\end{figure}

\textbf{Real-world results.} To evaluate physical deployment capability, we validate MARS on four hardware tasks (Fig.~\ref{framework}(h)). As illustrated in Fig.~\ref{realtest}(b), MARS consistently achieves optimal success rates across all benchmarks, outperforming flow matching. More importantly, this multimodal expressivity is delivered with extreme computational efficiency. As shown in Fig.~\ref{realtest}(d), MARS achieves a remarkably low inference latency ($\approx 5$~ms). This performance nearly matches A2A execution ($\approx 3$~ms) and is roughly $6\times$ faster than flow matching ($\approx 28$~ms), demonstrating its readiness for real-time robotic control. Consequently, both MARS and A2A produce noticeably smoother trajectories than flow matching, the vivid demonstrations of which are provided in the supplementary video.

As for multimodality, Fig.~\ref{realtest}(c) reveals that MARS preserves a superior modal balance, whereas the A2A baseline expectedly exhibits strict unimodality. Interestingly, diverging slightly from simulation outcomes, MARS marginally outperforms flow matching in three physical environments. Notably, when constrained by sparse demonstrations or overtrained checkpoints (see Fig.~\ref{appen_fm}), flow matching tends to drift toward unimodality, yet MARS robustly preserves its multimodal expressivity. 

We hypothesize that real-world data introduces more confounding factors, such as subtle variations in initial robot poses or visual backgrounds. Under data scarcity or excessive training iterations, these real-world nuisances exacerbate overfitting, which implicitly collapses mode diversity under the same evaluation initial setting. Thanks to its adaptive scheduling, MARS only maintains the necessary noise level, exhibiting enhanced resilience against overfitting compared to flow matching. Disentangling the precise scaling laws between real-world multimodality, demonstration volume, and training epochs remains a compelling avenue for future work.

\begin{figure}
	\centering
	\includegraphics[trim={0 24 0 0}, width=1\linewidth]{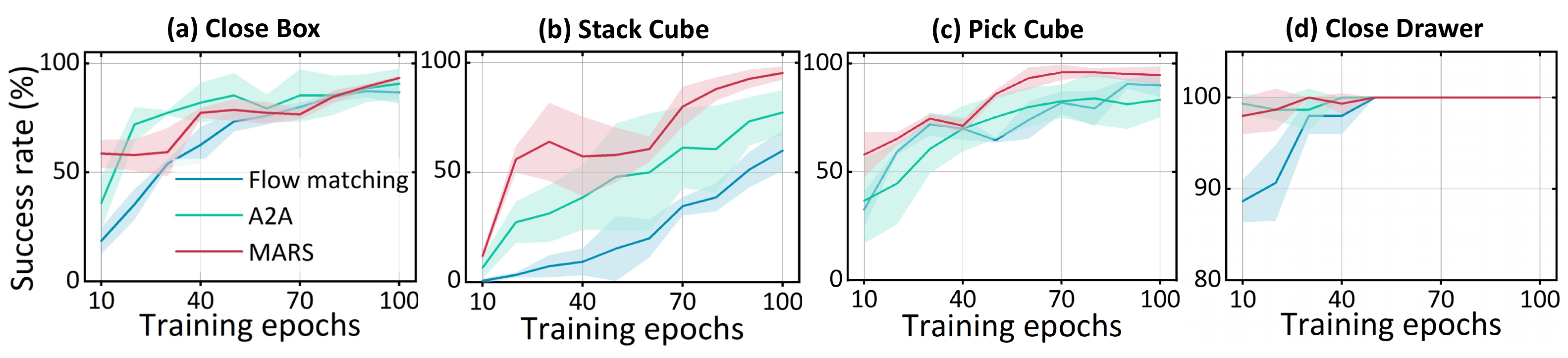}
	\caption{\textbf{Learning performance on strategically unimodal tasks.} (a) Close Box is from {RLBench}~\citep{james2020rlbench}. (b) Stack Cube and (c) Pick Cube are from {ManiSkill}~\citep{mu2021maniskill}. (d) Close Drawer is from {LIBERO}~\citep{liu2023libero}. The colored shaded area represents the standard deviations of $3$ random seeds.}
	\label{singlemodal}
\end{figure}

\subsection{Unimodal evaluation}

Next, MARS is carried out on 4 unimodal tasks. Fig.~\ref{singlemodal} reports the learning performance. As expected, MARS policy converges faster than flow matching in most cases, since unimodal tasks require less stochasticity and admit a more readily learnable observation-to-action mapping. More surprisingly, on tasks such as {Stack Cube} and {Pick Cube}, MARS policy converges even faster than A2A. We attribute this to two factors. On the one hand, MARS absorbs the subtle trajectory variations in expert demonstrations through stochastic initialization, whereas deterministic policies are forced to fit these variations as supervision noise, increasing gradient variance. On the other hand, the stochastic source acts as an input-side regularizer that smooths the loss landscape, an effect well-known to accelerate convergence in supervised learning.

We further evaluate the performance of the MARS policy against 8 SOTA baselines on the {Stack Cube} task across two distinct simulation platforms, \textit{MuJoCo} and \textit{IsaacSim}. As summarized in Tab.~\ref{tab-mujoco} and Tab.~\ref{tab-isaacsim}, MARS achieves a convergence rate comparable to the deterministic A2A baseline in both environments, while significantly outperforming other generative policy baselines.

\section{Conclusion}

We presented MARS policy, a flow-matching-based visuomotor policy that bridges the long-standing gap between deterministic policies (efficient but mode-averaging) and generative policies (expressive but slow). By exploiting the insight that multimodality in flow matching originates from stochastic initialization, MARS adaptively schedules the noise scale and inference steps according to the stochasticity required at each state. Experiments on 8 simulated and 4 real-world tasks across two kinds of platforms demonstrate that MARS achieves the learning efficiency of A2A and the multimodal expressivity of flow matching, which have so far been difficult to obtain simultaneously.

\section{Limitation}
The dispersion-based diversity loss, while effective in preserving multimodality, is not the most lightweight design. Both the BallTree construction and the per-iteration spread computation add overhead that scales with dataset size and neighbor count. Future work will explore more efficient alternatives, such as implicit dispersion regularizers or learnable diversity surrogates that avoid explicit neighbor queries altogether.

\bibliographystyle{assets/plainnat}
\bibliography{paper}

\clearpage
\newpage
\appendix

\section{Benchmarking on 2D Navigation Test}
\label{appen:2d_benchmark}
We benchmark the proposed MARS policy against a comprehensive set of baselines: \textit{Generative} paradigms, including DiT-based flow matching~\citep{flowmatching} and diffusion policy~\citep{dp}; \textit{Multimodal} baselines, such as IBC~\citep{florence2021implicit} and BET~\citep{shafiullah2022behavior}; \textit{Deterministic} policies, including A2A~\citep{jia2026action} and its stochastic counterpart Noised-A2A, VITA~\citep{gao2025vita}, and ACT~\citep{zhao2023learning}.

As shown in Fig. \ref{2d_benchmark}, while expert trajectories (a) demonstrate strategic multimodality, only stochastic models (b-g) can represent the underlying distribution. Notably, generative (b-d) achieves superior fidelity and cleaner trajectories compared to other methods. In contrast, deterministic baselines (h-j) are confined to a single mode, failing to capture the rich tactical variations present in the expert data. The Noised-A2A (g) further illustrates that slight noise injection is insufficient to recover the complex branching logic mastered by MARS.
    
\renewcommand*{\thefigure}{S1}
\begin{figure}
	\centering
	\includegraphics[width=0.5\linewidth]{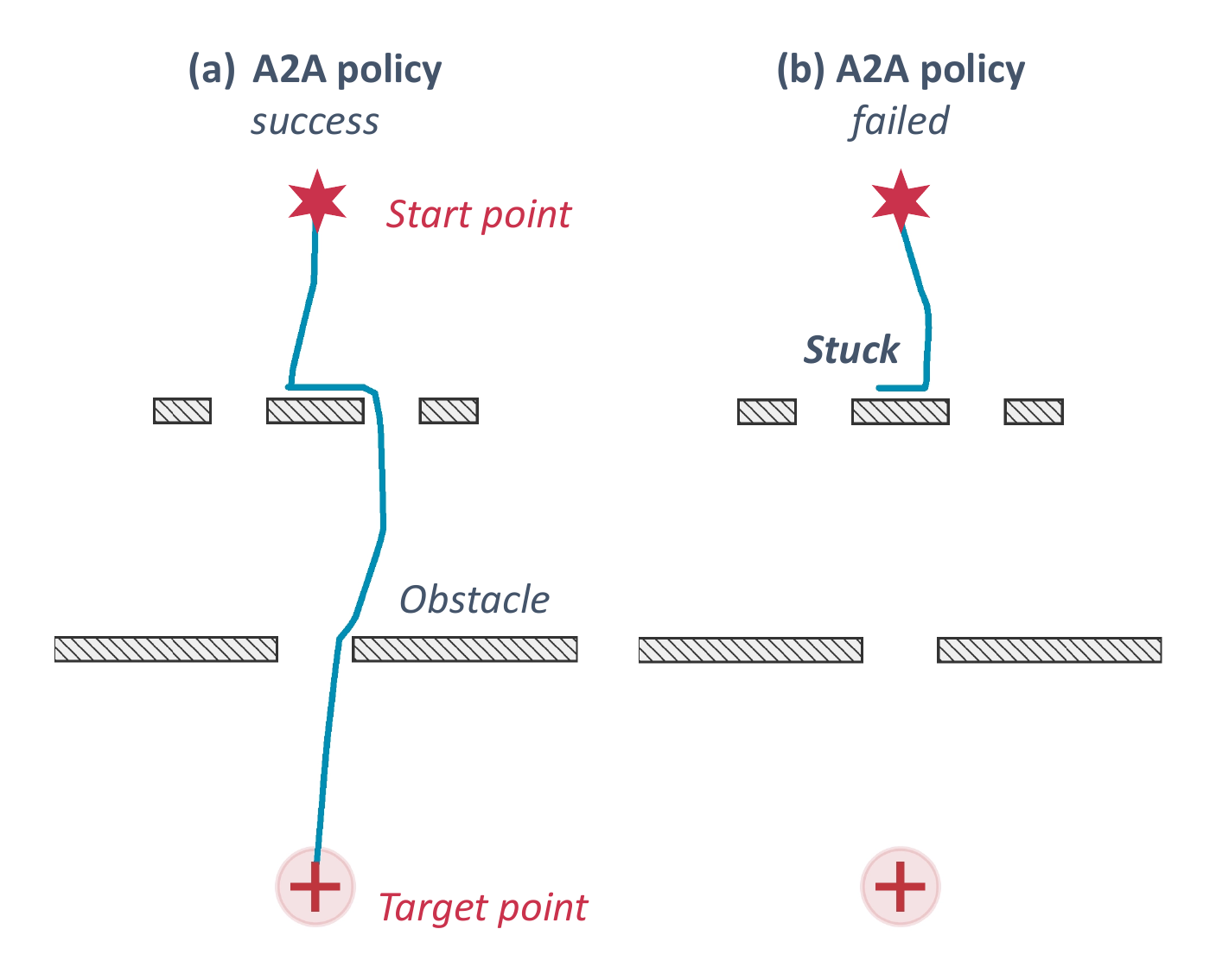}
	\caption{\textbf{A2A policy on 2D Navigation.} A2A occasionally reaches the target by collapsing to a single mode (a), but frequently gets stuck near an obstacle (b). Successful runs are largely attributable to training-time stochasticity.}
	\label{a2a_failed}
\end{figure}

\renewcommand*{\thefigure}{S2}
\begin{figure}
	\centering
	\includegraphics[width=1\linewidth]{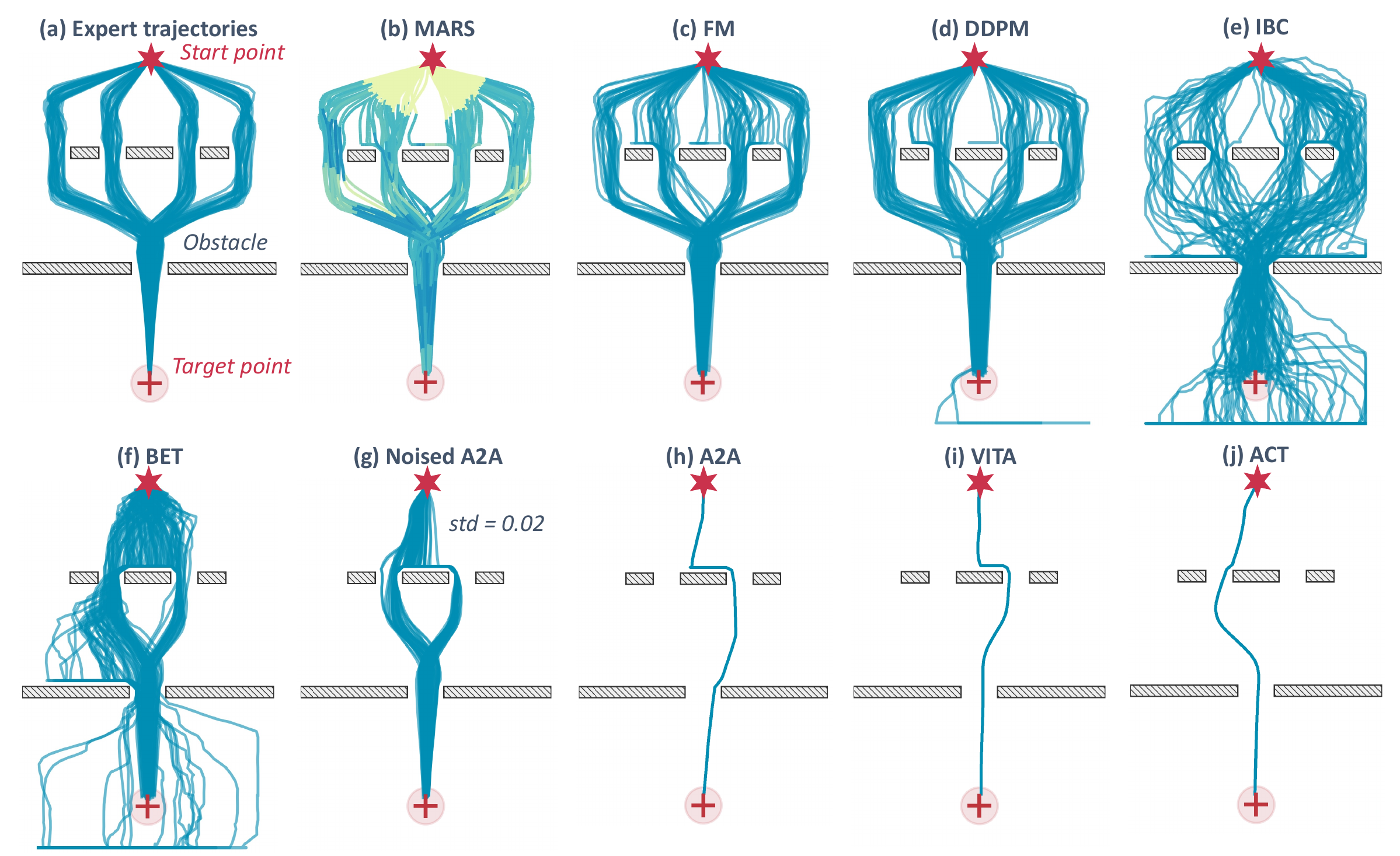}
	\caption{{Qualitative comparison of trajectory generation across different policy architectures.}}
	\label{2d_benchmark}
\end{figure}

\renewcommand*{\thefigure}{S3}
\begin{figure}
	\centering
	\includegraphics[width=0.6\linewidth]{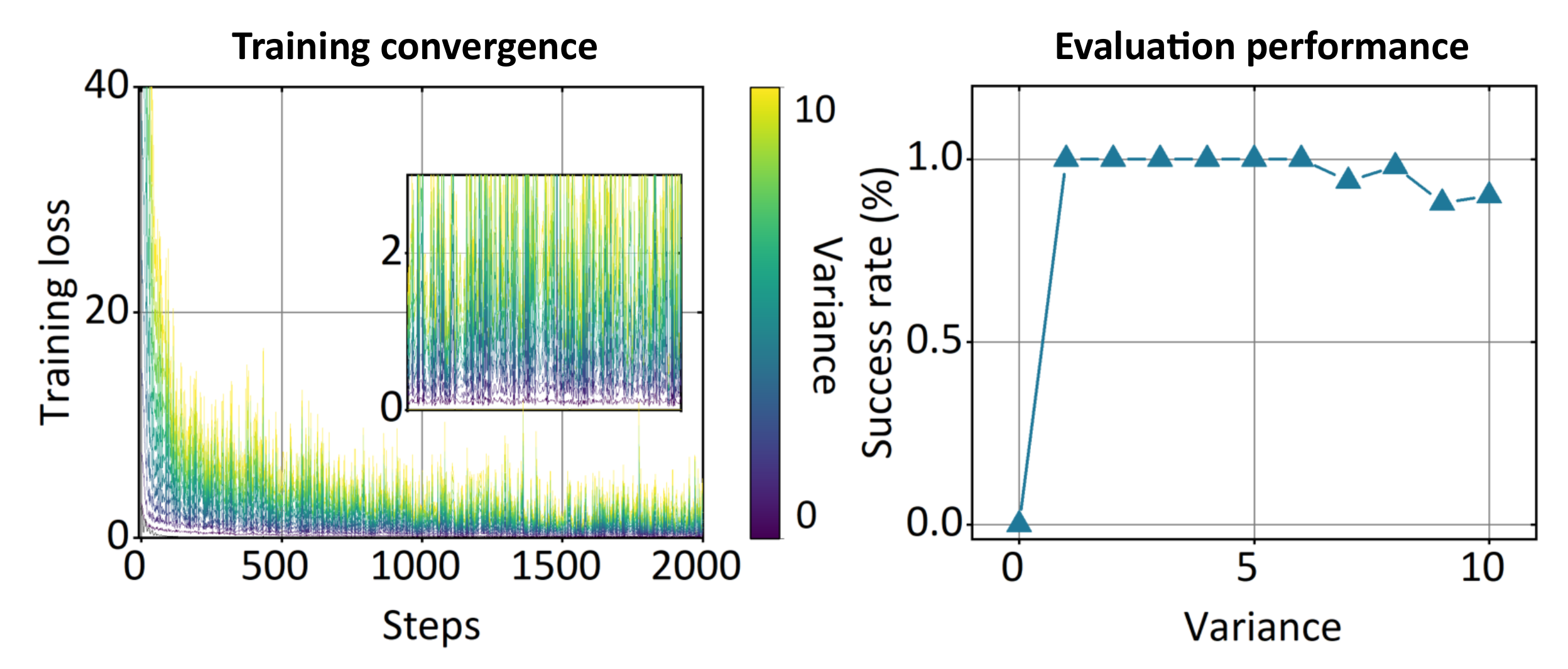}
	\caption{\textbf{Effect of initial flow variance on optimization and evaluation.} Settings: 30 training epochs, 100 demonstrations, and 50 evaluation rollouts. \textbf{Left:} Training loss curves under different initial variances (0-10) of the flow source distribution. Larger variance leads to slower and noisier convergence. \textbf{Right:} Evaluation success rate under different initial variances, showing that excessively large variance degrades success rate.}
	\label{loss_swap}
\end{figure}

\renewcommand*{\thefigure}{S4}
\begin{figure}
	\centering
	\includegraphics[width=0.5\linewidth]{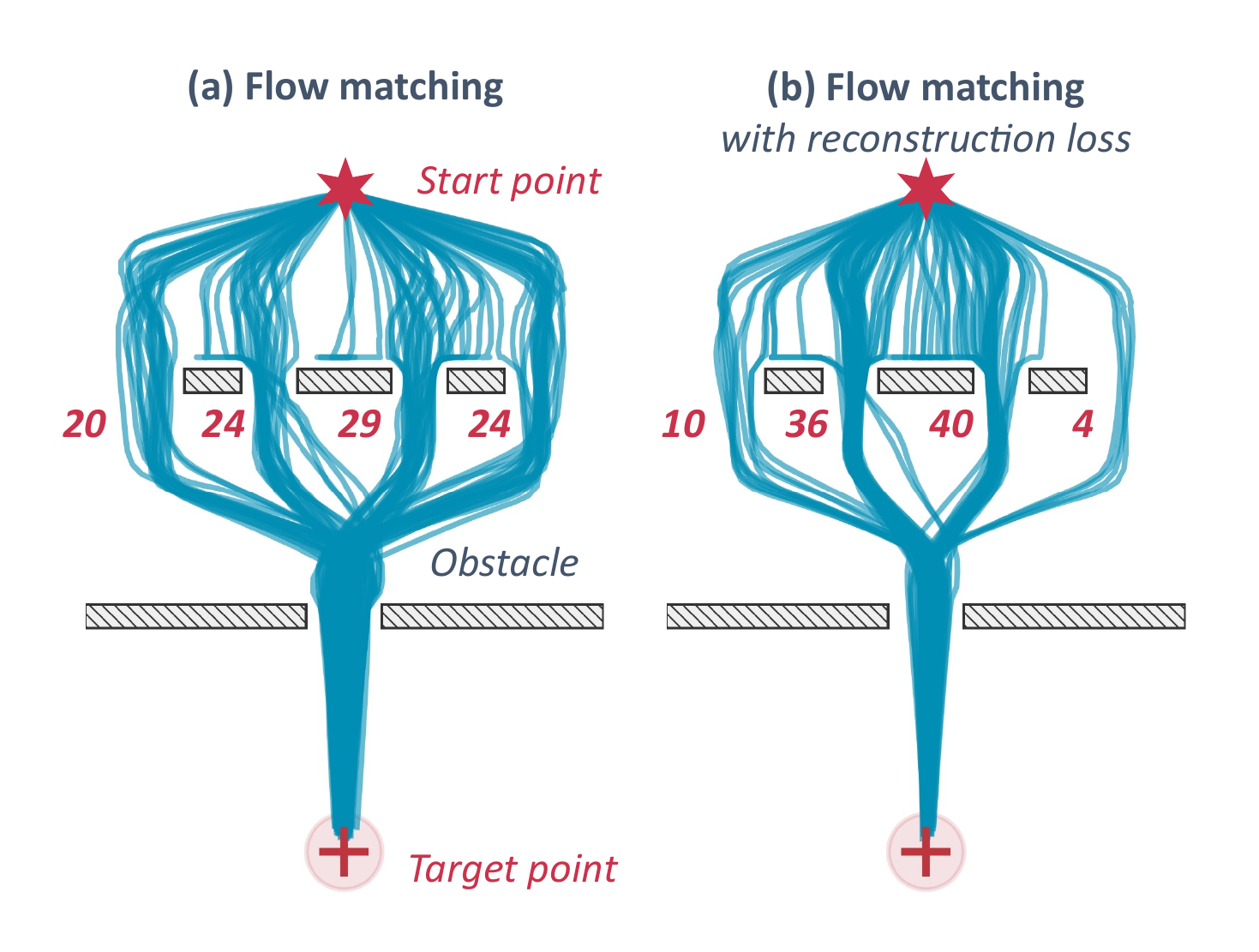}
	\caption{\textbf{Reconstruction loss compromises multimodal expressivity.} We visualize 100 rollouts of the flow matching policy on 2D Navigation. (a) Standard flow matching distributes trajectories across all four valid passages (counts: 20/24/29/24). (b) Adding a reconstruction loss with weight $\lambda_{\text{rec}} = 1$ biases the policy toward the two central passages (10/36/40/4), with the outer passages substantially under-explored. This motivates our per-dimension gating strategy that adaptively suppresses the reconstruction penalty based on recognized multimodal degree.}
	\label{fm_reconstruction}
\end{figure}

\section{Training and Evaluation Configurations}
\label{appen:training}

\subsection{Simulational Training and evaluation details}

\textbf{Training Details.}
 For unimodal benchmarks, we utilize $100$ expert demonstrations for {Stack Cube}, {Close Box}, and {Pick Cube}, while the {Close Drawer} task is trained with $50$ demonstrations. In multimodal scenarios, we employ $200$ demonstrations for the {2D Navigation} task, whereas all other tasks are trained using $100$ demonstrations. Additionally, the models presented in Fig. \ref{framework}, Fig. \ref{a2a_failed}, Fig. \ref{2d_benchmark}, and Fig.\ref{fm_reconstruction} are trained for a fixed duration of $50$ epochs. For the action space, the employed Franka uses joint commands in simulation.
 
 \textbf{ Evaluation Details.}
The \textit{2D Navigation} task is evaluated over $100$ trials using distinct random seeds to ensure statistical significance. For all other tasks, performance is reported based on the average of $50$ evaluation rollouts. All inference-time measurements in Fig. \ref{2d_benchmark} are conducted on a single NVIDIA $H200$ GPU ($141$ GB HBM3e) hosted on a node with dual Intel Xeon Gold 6530 processors. For every method, we report the mean and standard deviation of per-step inference time over $50$ evaluation rollouts. Moreover, the first five predictions per run are discarded as a warm-up.

\subsection{R1 Lite training and evaluation details}

\textbf{Hardware and Observation Setup.}
We evaluate Pick Cup and Pick Vegetable on a \textbf{Galaxea R1 Lite} bimanual robot. Only the right arm and right two-finger gripper are actuated during data collection and rollout. The policy observes two RGB views: a head-mounted third-person camera and a right wrist-mounted egocentric camera. For the action space, the R1 Lite uses joint commands in real-world experiments.

\textbf{Tasks and Demonstration Statistics.}
We evaluate on two contact-rich manipulation tasks. For each task, we deliberately collect a balanced set of demonstrations that contain \emph{two distinct interaction modes} so that we can probe the policy's ability to model multi-modal action distributions:

\begin{itemize}[leftmargin=1em, labelsep=0.4em, itemsep=0.25em, topsep=0.25em]
\item \textbf{Pick Cup.} The robot grasps a coffee mug from a tabletop. To induce multi-modality, we collect 100 demonstrations grasping the mug {by its rim} and 100 grasping it {by its handle}, yielding {200 demonstrations}. These geometrically distinct strategies start from the same initial states, making them challenging for unimodal policies. For this task, all three policies, A2A, flow matching, and MARS, are trained for 500 epochs.

\item \textbf{Pick Vegetable.}
The robot picks up a vegetable from a tabletop scene with a carrot, a mango, and a daikon arranged in sequence, where the carrot and daikon are placed on opposite sides of the mango. To induce multi-modality, we collect 100 demonstrations for picking the {carrot} and 100 for picking the {daikon}. Since either vegetable is a valid target, both modes are treated as successful task completions, yielding {200 demonstrations}. For this task, all three policies, A2A, flow matching, and MARS, are trained for 400 epochs. 

\end{itemize}
\textbf{Evaluation Details.}
For both Pick Cup and Pick Vegetable, we evaluate each trained policy with 10 rollout trials. In addition to task success, we record the executed action mode in each rollout, such as rim \textit{vs.} handle grasping for Pick Cup and carrot \textit{vs.} daikon picking for Pick Vegetable, to analyze whether each policy captures the intended multi-modal behavior.

\subsection{Franka training and evaluation details}

\textbf{Hardware and Observation Setup.}
We evaluate Push-T and Block Push using a \textbf{Franka Emika Research 3} (FR3) robotic arm. The scene is observed by one RGB-D Intel RealSense D455 camera, a fixed top-down camera providing a global view of the workspace. All human demonstrations are collected using a 3Dconnexion SpaceMouse via the HILSERL teleoperation framework~\citep{luo2025precise}. For the action space, the employed Franka uses end-effector states in real-world experiments.
\textbf{Tasks and Demonstration Statistics.}
Our evaluation uses two real-world manipulation tasks with action multimodality. We uniformly collect demonstrations across bimodal behavior modes, providing a basis for characterizing the multimodal structure of the action distribution.
\begin{itemize}[leftmargin=1em, labelsep=0.4em, itemsep=0.25em, topsep=0.25em]
\item \textbf{Push-T.} The end-effector pushes a T-shaped block into the target region from a fixed initial pose. The dataset contains {100 demonstrations} covering two pushing paths, corresponding to the upper and lower routes around the block, with 50 trials for each path. To reduce the influence of preceding motions during data collection, the robot is first moved to a randomly sampled point and then returned to the same start pose before each demonstration.

\item \textbf{Block Push.} The end-effector pushes one of two blocks to its corresponding target region from a shared initial pose.
The dataset contains {100 demonstrations} covering two pushing modes, pushing the orange block to its target region or pushing the purple block to its target region, with 50 trials for each mode.
This setup reflects action multimodality, as the same start states admit two valid object-selection choices.
\end{itemize}

\textbf{Training Details.}
For each Franka real-world task, we use $100$ demonstrations for training. All three policies, A2A, flow matching, and MARS, are trained with identical settings within each task.
Based on the convergence behavior of each task, we train the policies for 100 epochs on Push-T and 40 epochs on Block Push.
\textbf{Evaluation Details.}
For both Push-T and Block Push, we evaluate each trained policy with 20 rollout trials. 
Beyond task success, we also annotate the behavior mode executed in each rollout, including the upper or lower pushing route in Push-T and the selected block in Block Push.
This allows us to examine whether each policy preserves the demonstrated multimodal behaviors rather than converging to a single action pattern.
All inference-time measurements are  conducted on an NVIDIA RTX $5080$ GPU($16$ GB) and $128$ GB of system memory.

\subsection{Training and evaluation hyperparameters}

All training and evaluation hyperparameters of MARS policy are kept consistent across all experiments, as listed in Table~\ref{tab:parameters}.

\renewcommand*{\thetable}{S1}
\begin{table}[ht]
  \caption{Training hyperparameters.}
  \label{tab:parameters}
  \centering
  \begin{tabular}{lc}
    \toprule
    \textbf{Hyperparameters} & \textbf{Value} \\
    \midrule
    chunk size & 8 \\
    history horizon & 8 \\
    $K_{max}$  & 10 \\
    $m$  & 20 \\
    $\lambda_{\text{rec}}$   & 1 \\
    $\lambda_{\text{div}}$   & 1 \\
    Batch size           & 32 \\
    \bottomrule
  \end{tabular}
\end{table}

\section{Benchmarking on Cross-Sim Unimodal Tests}
\label{appen:Unimodal}
To validate the efficacy of the MARS policy, we conducted a comprehensive benchmark against eight SOTA baselines on the Stuck Cube task, utilizing both \textit{MuJoCo} and \textit{IsaacSim} as evaluation platforms. 
The compared baselines includes: DDPM-UNet~\citep{dp, ddpm}, DDPM-DiT~\citep{ddpm, Peebles_2023_ICCV}, DDIM-UNet~\citep{dp, song2020denoising}, FM-UNet~\citep{flowmatching}, FM-DiT~\citep{flowmatching, Peebles_2023_ICCV}, Score-UNet~\citep{song2020score}, ACT~\citep{zhao2023learning},  A2A \citep{jia2026action}. Key hyperparameters, including the chunk size, are standardized across all compared models to maintain experimental consistency. The evaluated results are summarized in  Tab. \ref{tab-mujoco} and Tab. \ref{tab-isaacsim}.

\renewcommand*{\thetable}{S2}
\begin{table*}[ht]
  \caption{Convergence rates of 9 algorithms on Stack Cube in \textit{MuJoCo} under different training epochs (100 demonstrations). The best results are highlighted in {bold}, while the second-best results are indicated with {underlines}.}
  \label{tab-mujoco}
  \vspace{3mm}
  \centering
  \small
  \setlength{\tabcolsep}{3.5pt} 
  \begin{tabular}{lccccccccccc}
    \toprule
    \textbf{Methods} & \textbf{Inference} & \textbf{10} & \textbf{20} & \textbf{30} & \textbf{40} & \textbf{50} & \textbf{60} & \textbf{70} & \textbf{80} & \textbf{90} & \textbf{100} \\
    & \textbf{Steps}  & (\%) & (\%) & (\%) & (\%) & (\%) & (\%) & (\%) & (\%) & (\%) & (\%) \\
    \midrule
    \textbf{MARS} & 1-10 & \textbf{14} & \textbf{50} & \textbf{78} & \textbf{74} & \textbf{68} & \underline{64} & \textbf{78} & \textbf{92} & \textbf{96} & \textbf{98} \\
    A2A                & 1 & \underline{12} & \underline{20} & \underline{44} & \underline{50} & \underline{60} & \textbf{72} & \underline{72} & \underline{74} & \underline{84} & \underline{86}\\
    FM-DiT              & 10 & 2 & 4 & 8 & 8 & 10 & 16 & 36 & 36 & 50 & 58\\
    DDPM-DiT            & 100 & 0 & 0 & 2 & 4 & 2 & 22 & 42 & 52 & 72 & 78\\
    FM-UNet             & 10 & 0 & 2 & 2 & 4 & 0 & 2 & 6 & 4 & 8 & 12\\
    DDPM-UNet           & 100 & 0 & 2 & 0 & 0 & 6 & 6 & 16 & 20 & 22 & 40\\
    DDIM-UNet           & 40 & 0 & 2 & 0 & 2 & 6 & 12 & 26 & 16 & 16 & 26\\
    Score-UNet          & 100 & 0 & 0 & 4 & 4 & 4 & 4 & 4 & 16 & 14 & 22\\
    ACT                 & 1 & 6 & 2 & 2  & 4 & 14 & 6 & 12 & 6 & 12 & 4 \\
    \bottomrule
  \end{tabular}
\end{table*}

\renewcommand*{\thetable}{S3}
\begin{table*}[ht]
  \caption{Convergence rates of 9 algorithms on Stack Cube in \textit{IsaacSim} under different training epochs (100 demonstrations). The best results are highlighted in {bold}, while the second-best results are indicated with {underlines}.}
  \label{tab-isaacsim}
  \vspace{3mm}
  \centering
  \small
  \setlength{\tabcolsep}{3.5pt} 
  \begin{tabular}{lccccccccccc}
    \toprule
    \textbf{Methods} & \textbf{Inference} & \textbf{10} & \textbf{20} & \textbf{30} & \textbf{40} & \textbf{50} & \textbf{60} & \textbf{70} & \textbf{80} & \textbf{90} & \textbf{100} \\
    & \textbf{Steps}  & (\%) & (\%) & (\%) & (\%) & (\%) & (\%) & (\%) & (\%) & (\%) & (\%) \\
    \midrule
    \textbf{MARS} & 1-10 & \textbf{10} & \textbf{50} & \underline{52} & \textbf{56} & \textbf{58} & \underline{60} & \textbf{74} & \textbf{80} & \textbf{92} & \textbf{90} \\
    A2A                & 1 & \underline{8} & \underline{44} & \textbf{54} & \textbf{56} & \underline{56} & \textbf{72} & \textbf{74} & \underline{74} & \underline{78} & \underline{84}\\
    FM-DiT               & 10 & 4 & 8 & 32 & 24 & 28 & 42 & 46 & 42 & 50 & 62\\
    FM-UNet             & 10 & 4 & 8 & 10 & 8 & 24 & 18 & 14 & 34 & 50 & 44\\
    DDPM-DiT             & 100 & 0 & 0 & 4 & 4 & 14 & 16 & 38 & 36 & 48 & 50\\
    DDPM-UNet            & 100 & 0 & 6 & 6 & 8 & 20 & 26 & 24 & 24 & 36 & 48\\
    DDIM-UNet            & 40 & 0 & 8 & 14 & 8 & 8 & 18 & 36 & 42 & 38 & 64\\
    Score-UNet          & 100 & 0 & 0 & 2 & 8 & 6 & 22 & 6 & 26 & 26 & 24\\
    ACT                  & 1 & 8 & 2 & 16 & 22 & 16 & 28 & 26 & 32 & 48 & 38\\
    \bottomrule
  \end{tabular}
\end{table*}

\section{Evaluation Metric of Multimodality}
\label{appen:metrics}

To quantify multimodal capability, we adopt a \textit{modal balance} metric $\gamma$ for the three bimodal tasks in Fig.~\ref{multimodal_sim}
\begin{equation}
    \gamma = \frac{2\min\{n_1, n_2\}}{n_1 + n_2},
\end{equation}
where $n_1$ and $n_2$ denote the number of successful rollouts assigned to mode 1 and mode 2, respectively. A value close to 1 indicates that the two modes are visited evenly, while a value approaching 0 reflects increasingly imbalanced mode coverage.

\renewcommand*{\thefigure}{S5}
\begin{figure}  
	\centering
	\includegraphics[width=0.9\linewidth]{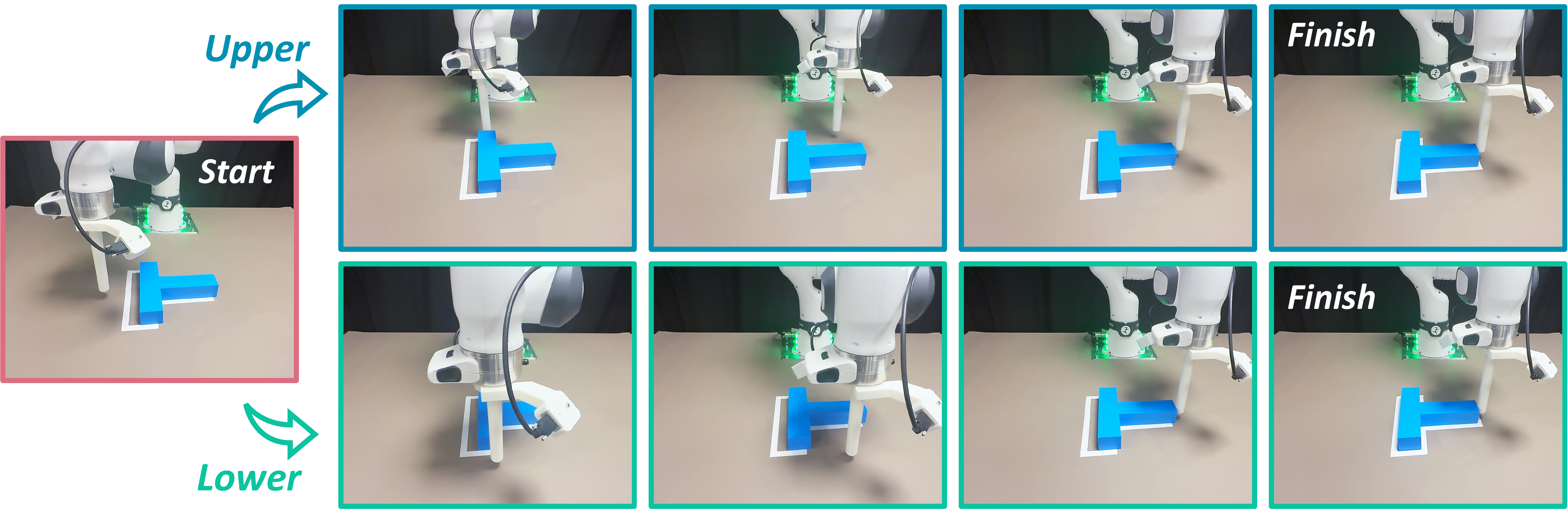}
	\caption{Snapshots of the Push-T task.}
	\label{pusht}
\end{figure}

\renewcommand*{\thefigure}{S6}
\begin{figure}
	\centering
	\includegraphics[width=0.5\linewidth]{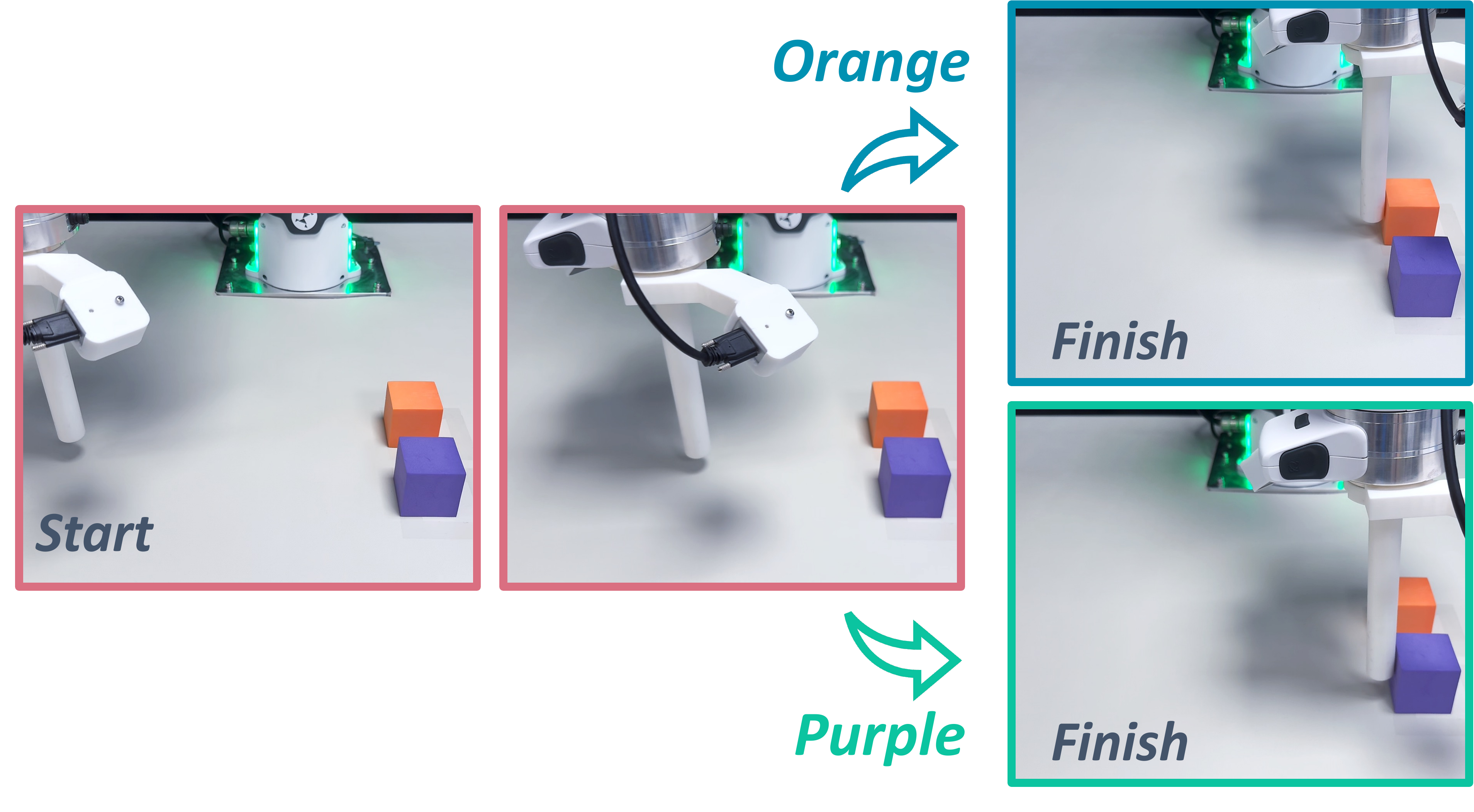}
	\caption{Snapshots of the Block Push task.}
	\label{pushcube}
\end{figure}

\renewcommand*{\thefigure}{S7}
\begin{figure}
	\centering
	\includegraphics[width=0.9\linewidth]{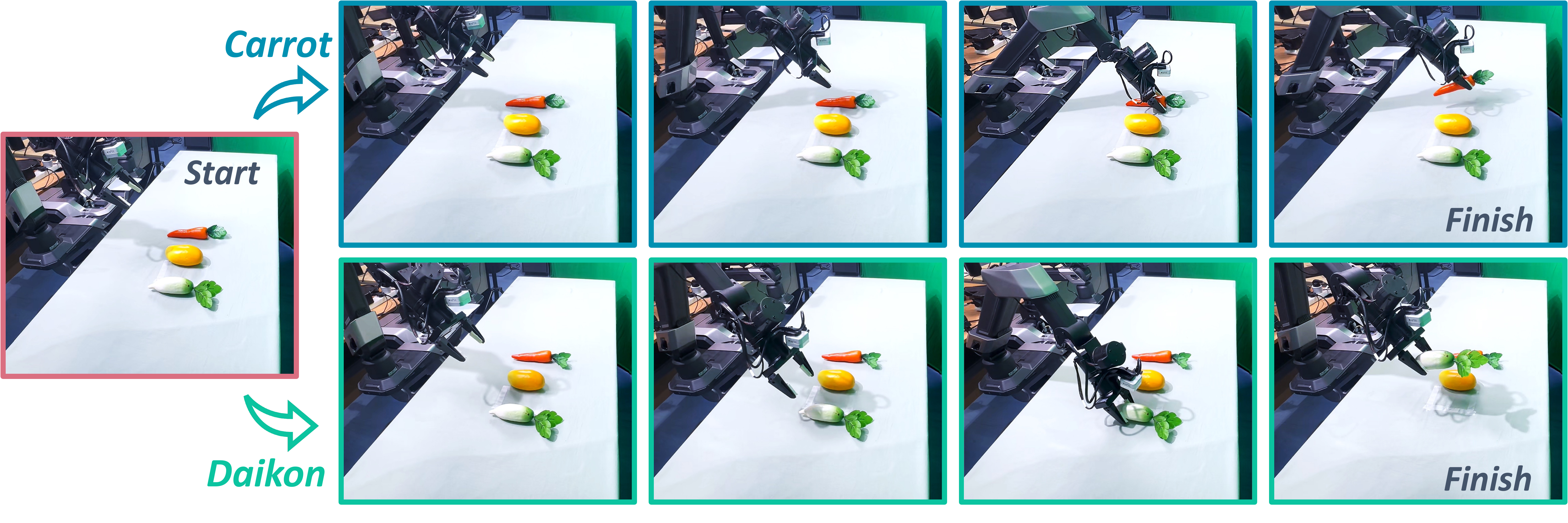}
	\caption{Snapshots of the Pick Vegetable task.}
	\label{pickvege}
\end{figure}

\renewcommand*{\thefigure}{S8}
\begin{figure}
	\centering
	\includegraphics[width=0.9\linewidth]{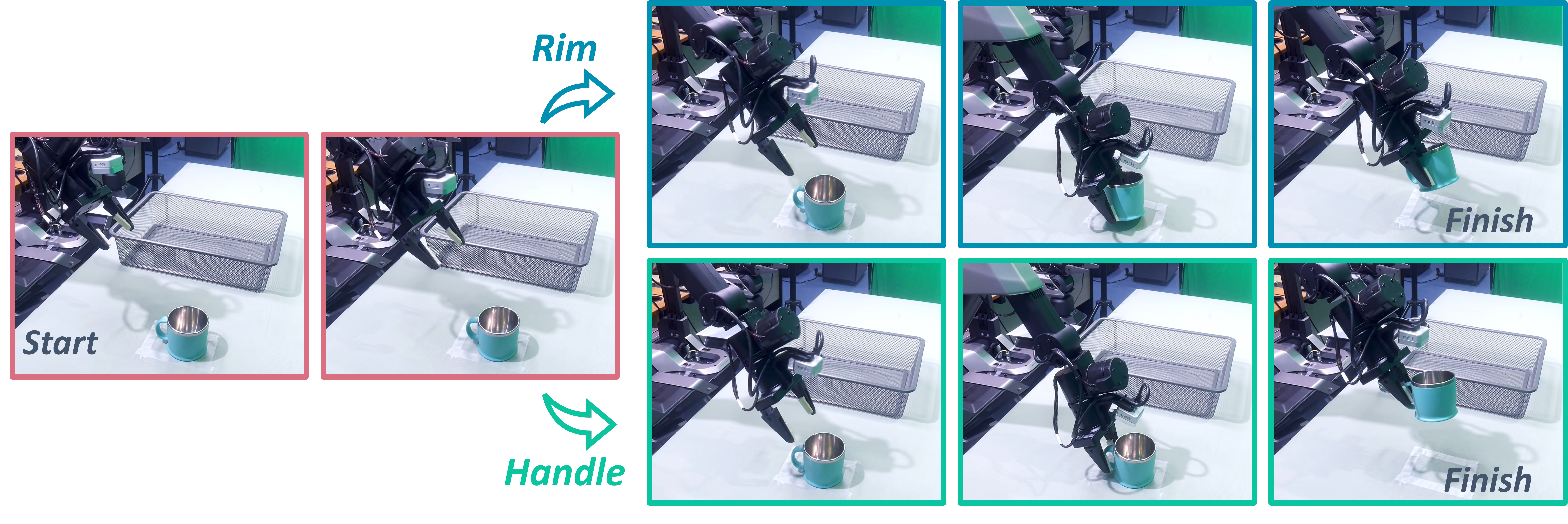}
	\caption{Snapshots of the Pick Cup task.}
	\label{pickcup}
\end{figure}

\renewcommand*{\thefigure}{S9}
\begin{figure}
	\centering
	\includegraphics[width=0.4\linewidth]{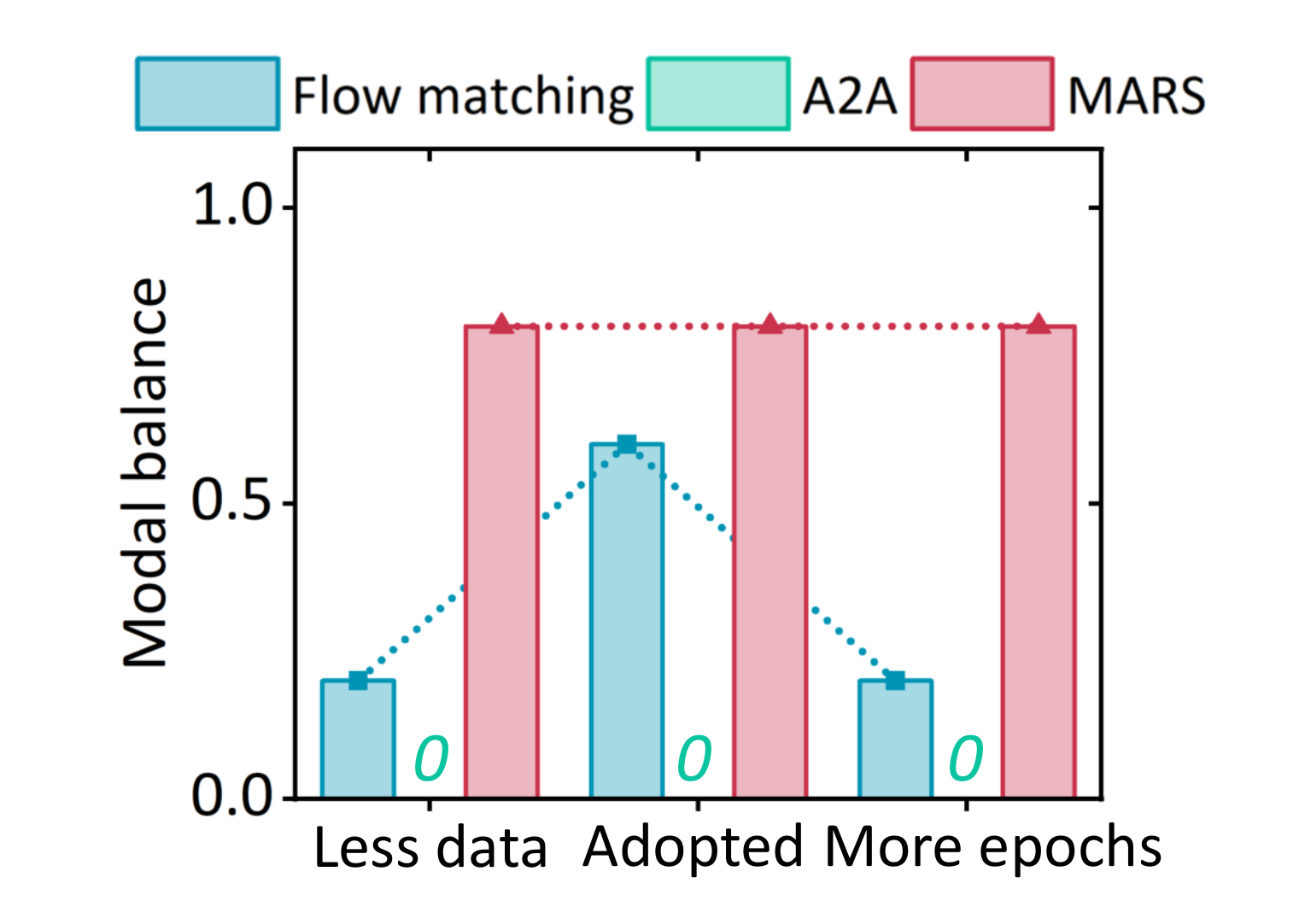}
	\caption{\textbf{Robustness analysis of modal balance under real-world data scarcity and overtraining (Pick Cup).} 
      The default configuration (``Adopted'') consists of 200 demonstrations and 400 training epochs. 
      MARS maintains a resilient and stable modal balance ($\gamma = 0.8$) across both data-restricted (50 demonstrations) and overtrained regimes (800 epochs). 
      In contrast, the generative baseline exhibits severe mode degradation. }
	\label{appen_fm}
\end{figure}

\end{document}